\documentclass[fleqn,12pt]{article}
\usepackage{graphicx}
\usepackage{latexsym}
\usepackage{a4}

\makeatletter
\newcounter{stelling}
\def\thestelling{\@arabic\c@stelling}

\newenvironment{example}{\refstepcounter{stelling} \begin{list}{{\bf Example \thestelling}}{} \item}{\end{list}}

\newenvironment{definition}{\refstepcounter{stelling} \begin{list}{{\bf Definition \thestelling}}{} \item}{\end{list}}

\newenvironment{ruledef}{\refstepcounter{stelling} \begin{list}{{\bf Rule \thestelling}}{} \item}{\end{list}}

\renewcommand{\models}{\mid\kern-2pt=}
\newcommand{\nm}{\mid\kern-2.1pt\sim}
\renewcommand{\vdash}{\mid\kern-5.8pt-}
\newcommand{\sat}{\mid\kern-2pt\equiv}

\title{Efficient Model Based Diagnosis}
\author{Nico Roos}
\date{}

\begin{document}
\maketitle

\noindent This paper has been published in {\it Intelligent
Systems Engineering}  (1993) 107-118.

\begin{abstract}
In this paper an efficient model based diagnostic process
is described for systems whose components possess a
causal relation between their inputs and their outputs.
In this diagnostic process, firstly, a set of focuses on
likely broken components is determined.
Secondly, for each focus the most informative probing
point within the focus can be determined.
Both these steps of the diagnostic process have a worst case
time complexity of ${\cal O}(n^2)$ where $n$ is the number
of components.
If the connectivity of the components is low, however, the
diagnostic process shows a linear time complexity.
It is also shown how the diagnostic process described can
be applied in dynamic systems and systems containing
loops.
When diagnosing dynamic systems it is possible to choose
between detecting intermitting faults or to improve the
diagnostic precision by assuming non-intermittency.
\end{abstract}

\section{Introduction}
The goal of Model Based Diagnosis is finding the cause of anomalies observed
in the behaviour of a (technical) system.
To reach this goal MBD, uses a model of the system to determine the
components that are responsible for the anomalies observed.
To determine these components, additional probing points may be selected.
The number of additional probing points, however, should be minimal.
Hence, we may not measure the output of every component in the system,
which would enable us to determine the diagnosis immediately.
Trying, on the other hand, to determine all the relevant information that is
implicitly available, i.e. the model of the system together with the
observation made, can be intractable \cite{Kle-91}.
To avoid this time complexity problem, several solution have been
proposed \cite{Kle-91,Bak-92}.
The solutions are all based on ignoring some of the information available.
This also holds for the diagnostic process described in this paper.
However, it also uses information is ignored by other diagnostic systems.

The diagnostic process described in this paper exploits the causal
relations between the inputs and the output of a component for
model based diagnosis.
Though the restriction to systems in which components possess a
causal relation between their inputs and their outputs,
seems to be a restriction on the applicability of the diagnostic
process described, it is not.
Components of information processing systems possess,
in general, causal relations between their inputs and their
outputs.
Also in physical systems, causality can be found.
Physical systems consisting of components of different fields,
such as mechanics, electromechanics, electromagnetism, physical
transport phenomena, chemical reaction kinetics, irreversible
thermodynamics, and others, can be modelled by using {\em bond graphs}
\cite{Dix-82}.
In a bond graph computational causality can be determined
between {\em all} components in algorithmic way.
As was argued in \cite{Top-91}, this computational causality corresponds
with physical causality if such a physical causality exists.
Hence, assuming causal relations between the inputs and
the output of a components
does not limit the applicability of a diagnostic process.

Using causality is not a new idea.
To determine a diagnosis, Davis \cite{Dav-84} has proposed the use of the causality for determining the
ancestor components of the faulty outputs of a system.
He uses the sets of ancestors of the faulty outputs to determine a
diagnosis using a {\em single-initial-cause heuristic}.
This heuristic implies that the intersection of these sets
has to be taken.
The resulting set is the set of possible candidates.
From this set Davis eliminates candidates using his
constraint suspension method.

The diagnostic process described in this paper is based on
a careful determination of the causal ancestors on which
the predicted value of a component's output {\em actually} depends.
Furthermore, it is based on analysing the influences of the observations
made on
the a posteriori fault probabilities given and the causal ancestors on which
the predicted output value of a component actually depends.
This has resulted in a diagnostic process in which, given a conflict
between a predicted and an observed output value of a component,
a {\it set of focuses} on the most likely broken components is determined.
Here, a focus is a set of components among which there is
probably one broken component.
The set of focuses is determined using a heuristic rule and can be viewed
as a generalisation of Davis's {\em single-initial-cause heuristic}.
Unlike the heuristics described by Chen and Srihari \cite{Che-89},
this rule is not based on
intuition, but on the analysis of a posteriori fault probabilities of the
components.

Given a focus, the most informative new probing  can be
determined.
The method described is based on the method described by
Bakker and Bourseau \cite{Bak-92}.
Here, the entropy of a probing point in a focus is determined
only using the focus and the set of components on which a predicted value of the probing point
actually depends.

The diagnostic process described is also applicable for dynamic
systems and systems containing loops.
In Section \ref{dynamic} it is described
how a set of focuses is determined in
a dynamic systems.
Depending on whether we want to be able to detect intermitting
faults, a different set of focuses can be determined.
Finally, in Section \ref{loops} diagnoses of systems containing
loops are discussed.

\section{The diagnostic process}
This section describes the diagnostic process.
First, a description of a system is introduced.
Here, a description of a system consists of the components, the connections
between components and the functions of the components.
Each component possesses only one output.
Since components with multiple outputs can be split into components
with only one output, this does not enforce a restriction on the
systems that can be modelled.
The system inputs must be represented by special components that
function as a source, and system outputs are a subset of the outputs of
the components.
There are no constraints on how the functions of the components are defined;
i.e. the functions
may be described by a composition of primitive functions, by logical
expressions or even by algorithms.

When making a diagnosis of a system, first we have to determine whether
there are anomalies.
For this we have to compute the output values of the components, starting with
known values of the system inputs and following the
causal direction.
When all relevant values of the system inputs are known, we
can predict the behaviour of the system and verify the observed behaviour.
Notice that unlike GDE \cite{Kle-87}, here the predictions are made in
only one direction through the system determined by the
causality of the components.
This will be called forward propagation here.
When an input value is predicted using the other input values
and the output value, this will be called backward propagation.

When the value of a component's output does not correspond with its predicted
value, we have a {\em conflict} and a diagnosis must be made.
Since here it is assumed that there is a causal relation between the inputs of a component and its output,
we know that one of the causal ancestors components
of the output for which we observed a conflict must be broken.
Because a component may function as a switch, the conflict observed
need not depend on all causal ancestors.
The set of components on which it {\em actually} depends is called the
{\em dependency set} of a component's output.

To guarantee that the predicted value of an output
depends on the corresponding dependency set,
a dependency set may not contain components whose output value have been measured.
Furthermore, each component in the dependency set of the output of a
component $c$
is either equal to $c$ or a component whose output is connected
to the input of a component in the dependency set.
Finally, notice that components that function as a switch cause
a special problem (confer \cite{Dav-88}).
The output value of a component that functions as a switch
need not depend on all the inputs of the component.
If the functions of the components are described by `if ..., then ...' rules,
the actual inputs on which a
predicted value depends can be determined.
However, there may exists hidden switches not described by
`if ..., then ...' rules.
Take for example a multiplier.
When one of its inputs is equal to zero, it functions as a switch.
And though this behaviour is implied by its function, unless stated
explicitly, it will not be recognised.
Fortunately, for the diagnostic process described below,
dependency sets that contain more
components than strictly necessary will only result in
some loss of efficiency.
\begin{definition} \label{depend}
Let $c$ be a component and let ${\it Dep}(c)$ be the
dependency set of the predicted output value of $c$.

${\it Dep}(c)$ is the smallest set of components satisfying the
following conditions:
\begin{itemize}
\setlength{\itemsep}{0pt}
\setlength{\parsep}{0pt}
\item
For no component $d$ such that: $d \in {\it Dep}(c)$ and $c \not= d$: $d$ has
been measured.
\item
For each component $d \in {\it Dep}(c)$ there holds: either $d = c$ or
for some $e \in {\it Dep}(c)$: the output of $d$ is connected to an
input of $e$.
\item
The output value of each component $d \in {\it Dep}(c)$ can be determine
from its measured input values and from the output values of
the components in ${\it Dep}(c)$ connected to its inputs.
An input value is measured if it is connected to an measured output
of some component.
\end{itemize}
\end{definition}

\begin{figure}[ht]
\hspace{1.6cm}
\includegraphics[scale=0.8]{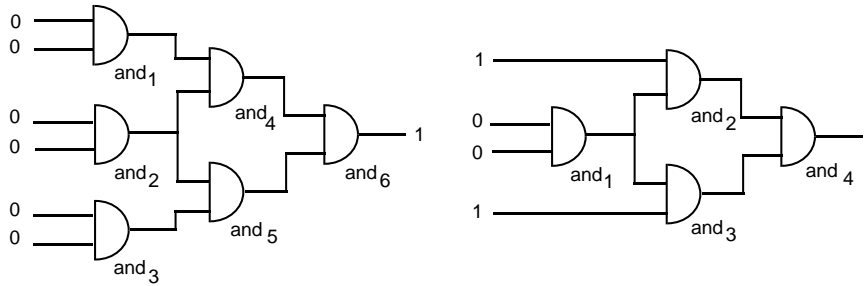}
\caption{multiple dependency sets} \label{bakker}
\end{figure}

If we describe that a multiplier functions as a switch when one
of its inputs is equal to zero, another problem arises.
When both inputs of a multiplier are equal to zero, the output is
determined by either one or the other input.
Hence, we get two dependency sets for one output.
Unfortunately, if multiple dependency sets for one output are possible,
the number of dependency sets for an output can become exponential
as is shown with the digital circuits in Figure \ref{bakker}.
To overcome this problem we can use the intersection of the dependency
sets of an output.
This resulting dependency set is called a {\em focused dependency set}.
\begin{definition}
Let $c$ be a component and let $Dep_1(c),...,Dep_n(c)$ be the dependency
sets of the output value of $c$.

Then the {\em focused dependency set} for $c$ is defined as:
\[ {\it Dep}^{\rm f}(c) = \bigcap \{ Dep_1(c),...,Dep_n(c) \}. \]
\end{definition}

The focused dependency set of a component $c$ defined above
contains the components that {\em can} influence the predicted
output value of $c$ if one of them is broken.
However, they need not influence the predicted value because
faults can be masked by other faults and by components whose
function is not {\em injective}.
Faults that are masked by other fault do not raise a problem as
will be shown in the next section.
However, components that can mask a fault because their
function is injective do raise a problem.
A comparator is an example of such a component.
These components make it impossible to assume that a focused
dependency set for a component $c$ is likely to contain
no broken component when the predicted output value of
$c$ is confirmed by an observation.
Therefore, a third kind of dependency set for a component,
called a {\em mask-free dependency set}, will be defined.
If exactly one of the components in this set for a component
$c$ is broken, then the output of $c$ is likely to incorrect.
\begin{definition} \label{mask_free}
Let ${\it Dep}^{\rm mf}(c)$ be a mask-free dependency set for
a component $c$.

${\it Dep}^{\rm mf}(c)$ is the largest subset of ${\it Dep}^{\rm f}(c)$
such that for each $d \in {\it Dep}^{\rm mf}(c)$ the probability that
the output value of $c$ will not change due to an incorrect output value
of $d$ is less than some reference value $\epsilon$.
\end{definition}

Both focused dependency sets an mask-free dependency sets can be
determined during forward propagation of predictions through the
system.
Each time an output value of a component $c$ is predicted given its
input values, we can also determine the focused dependency set of
this component given the focused dependency sets of the components
the outputs of which are connected to its inputs.
Let $d_1, ... ,d_m$ be the components the outputs of which satisfy the
following conditions:
\begin{itemize}
\item
The output of $d_i$ is connected to an input of the component $c$.
\item
The output value of $d_i$ has not been measured.
\end{itemize}
Furthermore, let $\Gamma_1,...,\Gamma_k $
be the minimal subsets of $\{ d_1, ... ,d_m \}$ needed to predict
the output value of $c$.
Then the focused dependency set of $c$ can be determined
in the following way: \[ {\it Dep}^{\rm f}(c) = \{ c \} \cup
(\bigcap \{ \Delta_1,...,\Delta_k \} ), \]
\noindent
where
$\Delta_i = \bigcup \{ {\it Dep}^{\rm f}(d_j) \mid d_j \in \Gamma_i \}$.
In a similar way the mask-free dependency set of $c$ can be
{\it approximated}.
Let $e_1, ... ,e_n$ be the components in $\{ d_1, ... ,d_m \}$ for which
there holds:
\begin{itemize}
\item
The probability that the output value of the component $c$
will not change due to an incorrect input value; i.e. an incorrect output value
of some component $e_i$, is less than some reference value $\epsilon$.
\end{itemize}
Then the mask-free dependency set of $c$ can be determined
in the following way: \[ {\it Dep}^{\rm mf}(c) = \{ c \} \cup
(\bigcap \{ \Sigma_1,...,\Sigma_k \} ), \]
\noindent
where $\Sigma_i = \bigcup \{ {\it Dep}^{\rm mf}(d_j) \mid d_j \in
(\Gamma_i \cap \{ e_1, ... ,e_n \} ) \}$.
Notice that this mask-free dependency set need not satisfy Definition
\ref{mask_free}. Definition \ref{mask_free} takes also into account
that sub-systems can mask a fault though none of its components
has this property.
In practise, however, these situations can be ignored.

Suppose that we have predicted the output values of the components and
have determined the corresponding focused dependency sets given
the initial measurements.
When the predicted output value {\em conflicts} with the measured output
value of a
component $c$, the {\em dependency set} ${\it Dep}(c)$ is called a {\em conflict set}.
Clearly, the dependency set contains at least one broken component.
If the measured output value {\em confirms} the predicted output value
of a component $c$, the {\em mask-free dependency} set
${\it Dep}^{\rm mf}(c)$ is called a {\em confirmation set}.
Since a confirmation set of $c$ contains the components
that are likely to influence the output value of $c$ if exactly
one of them is broken,
it contains components that are all correct,
or it contains a broken component whose fault is compensated
by others broken components in ${\it Dep}^{\rm f}(c)$.
\begin{definition}
If the measured output value of a component $c$ conflicts with
the predicted value, then the dependency set ${\it Dep}(c)$ is called
a conflict set $K(c)$ and the focused dependency set ${\it Dep}^{\rm f}(c)$
is called a focused conflict set $K^{\rm f}(c)$.

If the measured output value of a component $c$ confirms
the predicted value, then the mask-free dependency set
${\it Dep}^{\rm mf}(c)$ is called a confirmation set $B(c)$.
\end{definition}

If we have observed a conflict, each conflict set must contain at least one broken output of a component.
If such a conflict set contains only one component, we know that
this component is broken.
Hence the component is an element of the diagnosis.
If, however, it contains more than one component,
we must reduce the number of components by making new measurements.
By measuring an output value of a component $d$ in the conflict set
$K(c)$, the
conflict set is replaced by two new dependency sets, which are
proper subsets of the conflict set $K(c)$.
Given the measured output value of $d$ we will observe a new conflict for
$c$ or $d$.
In case of a double fault we will observe a conflict for both $c$ and $d$.
Hence, we will get either one or two new conflict sets, which are
proper subsets of $K(c)$.
This process is repeated till all conflict sets contain only
one component.
These components are the {\em diagnosis} of the anomalies observed.

\section{Focusing on likely broken components} \label{focus}
To minimise the number of additional measurements, we want to focus on
the most likely broken components.
Measuring their output values will give us more information
than measuring the output values of components of which it
is unlikely that they are broken.
The focus is determined by using the focused and the
mask-free dependency sets of the
measured output values of components.

Under the assumption that the components fail independently
of each other, we can estimate the a posteriori fault
probabilities with respect to the conflict and the confirmation
set observed.
Let $K_1,...,K_m$ be the conflict sets of the outputs
of the components $d_{1},...,d_{m}$,
let $B_1,...,B_n$ be the confirmation set of the
outputs of the components $e_1,...,e_n$,
and let $p_c$ be the a priori fault probability of a component $c$.
Then, for the a posteriori fault probability $p'_c$
of the component $c$ there holds:
\[ p'_c = p_c \cdot
\frac{Pr(K_1,...,K_m;B_1,...,B_n | c)}{Pr(K_1,...,K_m;B_1,...,B_n)}, \]
\noindent
where $Pr(K_1,...,K_m;B_1,...,B_n)$ and
$Pr(K_1,...,K_m;B_1,...,B_n | c)$
denote the probability that a conflict is observed for the outputs of
the components $d_{1},...,d_{m}$
and a confirmation for the output of $e_1,...,e_n$
(given that the component $c$
is malfunctioning).

Since $Pr(K_1,...,K_m;B_1,...,B_n )$ is the same for all
components $c$, it can be ignored.
Hence we only have to determine $Pr(K_1,...,K_m;B_1,...,B_n | c)$.
Here, two assumptions will be used.
Firstly, in general, the probability that one fault is compensated
by other faults is very small.
Therefore, we may assume that the components occurring in a
confirmation set are not broken.
So these components can be removed from the conflict sets.
Hence:
\[ Pr(K_1,...,K_m;B_1,...,B_n | c) \approx \left\{
\begin{array}{ll} 0 & \mbox{if } c \in (B_1 \cup ... \cup B_n) \\
Pr(\overline{K}_1,...,\overline{K}_m \mid c) &
\mbox{otherwise.} \end{array} \right.
\]
\noindent
where $\overline{K}_i = K_i - (B_1 \cup ... \cup B_n)$.

Secondly, assuming that the a priori fault probabilities of
the components lay in the interval $[a,b]$ where $a \gg b^2$, $Pr(\overline{K}_1,...,\overline{K}_m \mid c)$
can be approximated by the minimal number of components
such that each set $\overline{K}_i$ contains at least one
broken component.
This assumption can be justified by the observation that a priori
fault probabilities are very small.
These fault probabilities, which depend on the mean time between failure,
are of the magnitude of $10^{-10}$ or smaller.
Suppose that the largest fault probability is $10^{-10}$.
Then, according to the assumption the smallest fault probability should be
equal or larger than $10^{-18}$.
The reason for this requirement is that the chance of two
broken components with a priori fault probabilities of $10^{-10}$
is equal to the chance of one broken component with an a priori fault
probability of $10^{-20}$.
If the a priori fault probabilities are sufficiently small, in general, the
assumption will hold.

Given this assumption the probability
$Pr(\overline{K}_1,...,\overline{K}_m \mid c)$ depends on the minimal number of components needed to explain the conflicts corresponding with
$\overline{K}_1,...,\overline{K}_m$ given that $c$ is broken.
Hence, it depends of the number of components in a minimal hitting set
of $\{ \overline{K}_i \mid c \not\in \overline{K}_i \}$.
Unfortunately, the determination of a minimal hitting set is
an NP-hard problem.
Therefore an approximation will be used.

Observe that for each conflict set $K_i$ there is exactly one
corresponding focused conflict set $K^{\rm f}_j \subseteq K_i$.
Clearly, if each focused conflict set in
$\{ \overline{K}^{\rm f}_j \mid c \not\in \overline{K}^{\rm f}_j \}$
contains one broken component, so does each conflict set in
$\{ \overline{K}_i \mid c \not\in \overline{K}_i \}$.
Because for the number of components $n$ in a minimal
hitting set of $\{ \overline{K}_i \mid c \not\in \overline{K}_i \}$
there holds:
$n \leq | \{ \overline{K}^{\rm f}_j \mid
c \not\in \overline{K}^{\rm f}_j \} | \leq
| \{ \overline{K}_i \mid c \not\in \overline{K}_i \} | $,
we can use the size of the set
$\{ \overline{K}^{\rm f}_j \mid c \not\in \overline{K}^{\rm f}_j \}$
as an approximation.
Hence, the following intuitive focusing rule can be formulated.
\begin{ruledef} \label{r1}
Given the components in a focused conflict set, focus on the
components occurring in the largest number of other focused conflict sets but in no confirmation set.
\end{ruledef}

Because this rule is based on determining an upper bound for the number of
components in the minimal hitting set of
$\{ \overline{K}_i \mid c \not\in \overline{K}_i \}$,
the focus determined by the rule can be incorrect in cases of multiple
faults.
Experiments, however, indicate that this problem will not often arise.

By applying the focusing rule, we can determine a set of focuses, each
probably containing one broken component.
This set of focuses can contain non minimal elements with
respect to the inclusion relation.
Because if a focus contains a broken component, every focus
that is a superset will also contain a broken component,
we may remove the non minimal elements.

\begin{figure}[ht]
\hspace{4.5cm}
\includegraphics[scale=0.5]{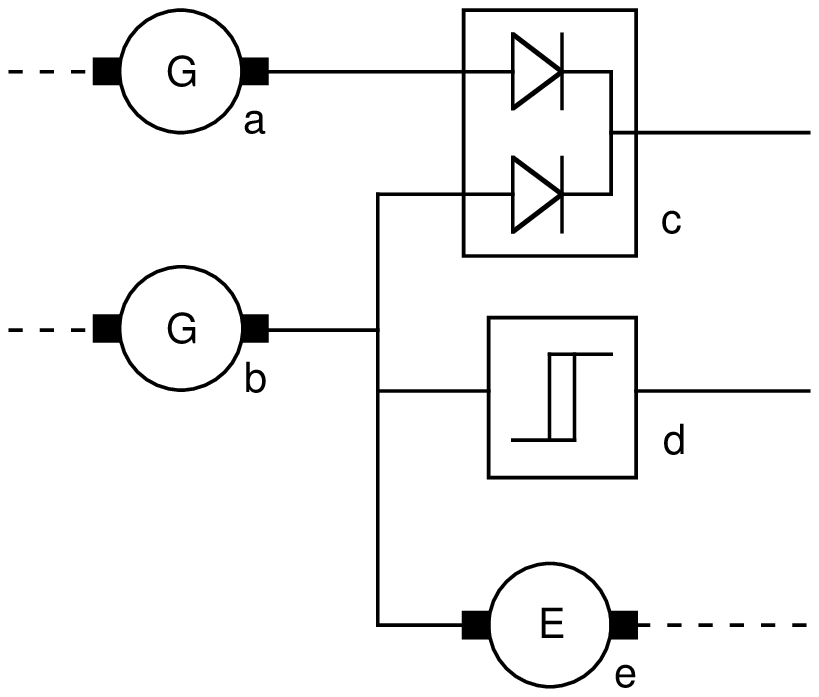}
\caption{} \label{demo}
\end{figure}
\begin{example}
To illustrate the application of the focusing rule, consider
the system in Figure \ref{demo} consisting of two DC power generators
$a$ and $b$, a diode bridge $c$, a devise $d$ indicating whether generator
$b$ can supply power and an electrical engine $e$.
If both generators $a$ and $b$ are in use, we can determine the following
dependency sets.
\begin{list}{}{} \item
${\it Dep}_1(c) = \{ a,c \}$ \\
${\it Dep}_2(c) = \{ b,c \}$ \\
${\it Dep}^{\rm f}(c) = {\it Dep}^{\rm mf}(c) = \{ c \}$ \\
${\it Dep}(d) = {\it Dep}^{\rm f}(d) = \{ b,d \}$ \\
${\it Dep}^{\rm mf}(d) = \{ d \}$ \\
${\it Dep}(e) = {\it Dep}^{\rm f}(e) = {\it Dep}^{\rm mf}(e) = \{ b,d \}$
\end{list}
Notice that electrical power will be available on the output of $c$
if one of the generators
can supply it, and that the signal of $d$ cannot indicate that
voltage of generator $b$ is too low.

Now suppose that the engine $e$ is not working property.
If the output values of $c$ and $d$ confirm their predicted values,
we can determine one focus $\{ b,e \}$.
If also the output value of $d$ conflicts with its predicted value, the focus
will be $\{ b \}$.
Finally, if output values of both $c$ and $d$ conflict with their
predicted values, we can determine two focuses: $\{ b\}$ and $\{ c \}$.
Clearly, the focus $\{ c \}$ is too narrow and should have been equal
to $\{ a,c \}$.
\end{example}

Above it was assumed that the probability that a fault is compensated
by other faults is very small.
In general, this assumption is valid when the domain of a component's
output values is sufficiently large.
For systems such as digital circuits, however, the assumption does not hold.
For such systems we must consider the possibility that less
broken components are needed to explain the observations made
when one fault is compensated by other faults.
Hence, to determine the relative likelihood of
$Pr(K_1,...,K_m;B_1,...,B_n | c)$ we must determine the
minimal number of components such that each conflict set
contains at least one broken component and each confirmation set
contains either zero or at least two broken components
whose fault compensate each other.
Because this problem is even harder than determining a
minimal hitting set of the conflict sets, again an approximation
will be used.
This approximation is based on the assumption that the mask-free
dependency set of a component is equal to its focused dependency set.

If the component $c$, which is assumed to be broken,
is a member of a confirmation set,
we need at least one other broken component in the
confirmation set
to compensate the fault caused by $c$.
Hence, if $c$ is broken, then, under the assumption that the mask-free
dependency set of a component is equal to its focused dependency set,
the confirmation set becomes focused conflict set in which
there is a broken component not equal to $c$.
Furthermore, the fault caused by $c$ cannot be compensated
by a component that belongs to the original conflict sets.
We would not observe conflicts caused by the component $c$
if the fault of $c$ is compensated by other components in these conflict
sets.
Hence, if $c$ is broken, then, under the assumption that the mask-free
dependency set of a component is equal to its focused dependency set,
the focused conflict sets containing $c$
become confirmation sets and the confirmation sets containing $c$
become focused conflict set.
This implies that each confirmation set $B_j$ with $c \in B_j$
must contain at least one broken component $d$ for which there
holds: if $c \in K^{\rm f}_i$, then $d \not\in K^{\rm f}_i$.
Therefore, for any component $c$,
$| \{ K^{\rm f}_k \mid c \not\in K^{\rm f}_k \}
\cup \{ B_j \mid c \in B_j \} |$ is an
upper bound for the minimal number of additional broken
components.
Given this result we can determine a focus using the
following rule.
\begin{ruledef} \label{r2}
Given the components in a focused conflict set, focus on the
components that maximise difference between the  number of focused conflict sets of which they are a member and the
number of confirmation sets of which they are a member.
\end{ruledef}

Notice that the set of focuses determine by this rule
need not be complete.
If a component $e$ in a focus is also an element of some
confirmation set, their must be an other broken component
in the confirmation set.
No focus containing this component is determined because
such a focus depends on the component $e$.
By assuming that $e$ is broken we can determine this focus
in the following way.
Let $\Delta = \bigcup \{ K^{\rm f}_k \mid e \in K^{\rm f}_k \}$.
Then we can the determine the focus by only
using the conflict sets
$\{ K^{\rm f}_k \mid e \not\in K^{\rm f}_k \} \cup
\{ B_j - \Delta \mid e \in B_j \}$.

Using one of the two rules above we can determine
likely minimal candidate diagnoses (Rule \ref{r1}) or
the likely minimal partial candidate diagnoses (Rule \ref{r2}).
Because focuses need not always be mutually exclusive,
these candidate diagnoses must be determined by a hitting
set.
Fortunately, there is no need for determining the actual candidate diagnoses.
The set of focuses determined by either rule suffice for the determination
of a new probing point as will be shown in the next section.

\begin{figure}[ht]
\hspace{2.1cm}
\includegraphics{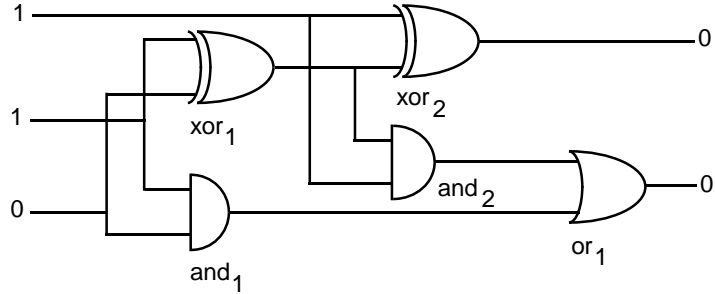}

\caption{full-adder} \label{adder}
\end{figure}
\begin{example}
Consider the full-adder in figure \ref{adder}.
For this full-adder we can predict the component's output values
and we can determine the component's focused dependency sets.
Notice that for digital circuits a mask-free dependency set
is equal to a focused dependency set.
\begin{tabbing}
mmm\=com\=ponentmm\=output\=mvaluemm\=mm\=\kill
\>component\>\>output value\>\>focused dependency set \\
\>\>$and_1$ \>\>0 \> $\{and_1\} $ \\
\>\>$xor_1$ \>\>1 \> $\{xor_1 \} $ \\
\>\>$and_2$ \>\>1 \> $\{and_2,xor_1\} $ \\
\>\>$xor_2$ \>\>0 \> $\{xor_1,xor_2\} $ \\
\>\>$or_1$ \>\>1 \> $\{or_1,and_2,xor_1\}$ \\
\end{tabbing}
Since we have a conflict between the measured and the predicted output
value of $or_1$, there must be a broken component in:
${\it Dep}^{\rm f}(or_1) = \{ or_1,and_2,xor_1 \}$.
From this set we can select a focus by applying the second heuristic rule.
Since the output of $xor_2$ is confirmed, the focus becomes:
$\{and_2, or_1\}$.
Applying the divide and conquer strategy the output to be
measured next is the output of component $and_2$.
\end{example}

The application of both focusing rules introduced above requires, in the
worst case a time complexity of ${\cal O}(n^2)$ where $n$ is the
number of components.
Since the prediction of the behaviour of the system and the
determination of the focused dependency sets can be done in linear time,
the whole diagnostic process has a worst case time complexity
of ${\cal O}(n^2)$.
However, if the connectivity of the components is low, the time
complexity can become linear.

\section{Probing points} \label{entropy}
Because a focus may contain more than one component,
we need to select a probing point.
This, for example, can be done by applying a divide
and conquer strategy.
So, the output of a component $c$ to be measured next
should be chosen such that:
\[ | F - {\it Dep}^{\rm f}(c) | \approx \frac{1}{2} \cdot | F |, \]
\noindent
where $F$ is the focus.
Notice that there is no need for choosing as a probing point the output
of a component in the focus $F$.

Instead of a divide and conquer strategy, we can also use the
entropy of a probing point.
As was shown by Bakker and Bourseau \cite{Bak-92},
a new probing point can be determined without generating the
candidate diagnoses, by only using the conflict sets.
Because for applying the focusing method, it suffices to know
whether a measurement confirms or conflicts with the predicted
value, the method for given probing advise can be simplified.
Unfortunately, this simplification is only applicable if the mask-free dependency set of
a candidate probing point is equal to the
corresponding focused dependency set.
This condition is necessary for two reasons.
Firstly, it is necessary to determine the probability that we will
measure a confirmation of the predicted value of a
candidate probing point.
Secondly, it is necessary to determine what the new focus is
going to be only using the current focus and the dependency sets
of a candidate probing point.
Examples of systems where the mask-free dependency set
is equal to the focused dependency set are: digital circuits
and energy conserving systems.

Since we only distinguish two situation; a measurement
confirms or conflicts with the predicted value, the determination
of the probing point with the highest entropy
can be approximated by selecting the probing point that splits
the a priori probability of having at least one broken component
in the focus in half.
I.e. we should measure the output of that component $c$ for which
there holds that $Pr(b(c) \mid F) \approx \frac{1}{2}$,
where $F$ denotes that
there is at least one broken component in the focus $F$ and
$b(c)$ denotes that
the measurement confirms the predicted output value of the component $c$.
When the measurement confirms the predicted output value
of a component $c$, its focused dependency set ${\it Dep}^{\rm f}(c)$ is not
likely to contain a broken component.
Hence, the new focus would become $F' = F - {\it Dep}^{\rm f}(c)$.
Therefore, if $Pr(X)$ denotes the a priori probability that a set
of components $X$ contains at least one broken component, then:
\[ Pr(b(c) \mid F) = \frac{Pr(F')}{Pr(F)}. \]

The probability that a set of components contains at least one
broken component can be determined in the following way.
\[ Pr({X}) = \left\{ \begin{array}{ll}
1 - \prod_{c \in X} (1 - p_c) & \mbox{if } X \not= \emptyset \\
0 & {\rm otherwise,} \end{array} \right. \]
\noindent
where $p_c$ denotes the a priori fault probability of a component $c$.
The determination of a probability $Pr(X)$ can be done in
linear time with respect to the number of components in $X$.
Since both a set $X$ and a focus can contain at most
$n$ components,
the determination a new probing point has a worst
case time complexity of ${\cal O}(n^2)$.

As was mentioned above the probing advise cannot be used in case
the mask-free dependency set of a potential probing point is not
equal to its focused dependency set.
In such cases it is still possible that a component in
$Def^{\rm f}(c) - Def^{\rm mf}(c)$ is broken after measuring an
output value for $c$ that confirms its predicted value.
Hence, the following inequality holds for $Pr(b(c) \mid F)$.
\[ \frac{Pr(F')}{Pr(F)} \leq Pr(b(c) \mid F) \leq \frac{Pr(F'')}{Pr(F)}, \]
\noindent
where $F' = F - Def^{\rm f}(c)$ and $F'' = F - Def^{\rm mf}(c)$.
This inequality can be used instead to select more roughly a
new probing point.

\section{Focusing in dynamic systems} \label{dynamic}
The focusing method described in Section \ref{focus} can
also be applied in MBD of dynamic systems.
To be able to apply it, we need to make samples of a number
of probing points of the system, and compare the measured
values with the values predicted by simulating the behaviour
of the system.
Again we will get focused conflict sets and confirmation sets.
Here, however, these conflict and confirmation sets depend
on the time points on which we sampled the system.
Therefore, the definition of a dependency set has to be modified.
\begin{definition}
Let $c$ be a component, let $t$ be a time point and let ${\it Dep}(c,t)$
be the dependency set of the
predicted output value of $c$ on time point $t$.

${\it Dep}(c,t)$ is the smallest set satisfying the following conditions:
\begin{itemize}
\item
Fore each $\langle d,t' \rangle \in {\it Dep}(c,t)$ there holds: $t' \leq t$.
\item
For no component $d$ such that: $\langle d,t' \rangle \in {\it Dep}(c,t)$ and
$d \not= c$: $d$ has been measured on time point  $t'$.
\item
For each component $d$ such that: $\langle d,t' \rangle \in {\it Dep}(c,t)$
there holds: either $d = c$ or
for some $\langle e,t'' \rangle \in {\it Dep}(c,t)$: the output of
$d$ is connected to
an input of $e$ and $t' \leq t''$.
\item
The output value of each component $\langle d,t' \rangle \in {\it Dep}(c,t)$
can be determine from its measured input values and from the output
values of the components in ${\it Dep}(c,t)$ connected to its inputs.
\end{itemize}
\end{definition}
\begin{definition}
Let ${\it Dep}^{\rm mf}(c,t)$ be a mask-free dependency set for
a component $c$ on time point $t$.

${\it Dep}^{\rm mf}(c,t)$ is the largest subset of ${\it Dep}^{\rm f}(c,t)$
such that for each $\langle d,t' \rangle \in {\it Dep}^{\rm mf}(c)$
the probability that the output value of $c$ will not change on time point
$t$ due to an incorrect output value
of $d$ on time point $t'$ is less than some reference value $\epsilon$.
\end{definition}
Notice that these definitions enables us to denote the time points on
which a component must have functioned correctly to observe
the predicted value.

Below a reformulation of Rule \ref{r1} for dynamic systems
will be given.
In this new version of Rule \ref{r1}, components that occur in some
confirmation set are replaced by {\em cancelled} components.
So we focus on the components occur in the largest number of
conflict set and which are not cancelled.
Whether a component is cancelled depends on whether we want to
be able to detect intermitting faults.
If  intermitting faults do not occur, a
component $c$ that is broken on time point $t$,  causes a fault in every
time point $t' \geq t$.
So for no time point $t' \geq t$: $\langle c, t' \rangle$ is an element
of some confirmation set.
Therefore, if $\langle c, t_1 \rangle \in K^{\rm f}(d,t_2)$
with $t \leq t_1$ and if there exists a confirmation set $B(e,t_3)$
such that: $\langle c, t_4 \rangle \in B(e,t_3)$ and
$t_1 \leq t_4$, then the component $c$ is cancelled for the
focused conflict set $K^{\rm f}(d,t_2)$.

If intermitting faults do occur, we can detect them by cancelling the components $c$
for a conflict set $K^{\rm f}(d,t_2)$ if for every
$\langle c, t_1 \rangle \in K^{\rm f}(d,t_2)$ there exits
a confirmation set $B(e,t_3)$
such that: $\langle c, t_1 \rangle \in B(e,t_3)$.
Notice that we cannot cancel a component for a focused conflict
set if the time point of the component in the focused conflict set
does not correspond with the time point in the confirmation set.
Because intermitting faults are possible the component can cause
the conflict observed.
\begin{definition}
If a component $c$ does not possess intermitting faults,
this component $c$ is cancelled
for a focused conflict set $K^{\rm f}(d,t_2)$ if  for each
$\langle c, t_1 \rangle \in K^{\rm f}(d,t_2)$
there exists a confirmation set $B(e,t_3)$
such that: $\langle c, t_4 \rangle \in B(e,t_3)$ and
$t_1 \leq t_4$.

If a component $c$ can possess intermitting faults,
this component $c$ is cancelled
for a focused conflict set $K^{\rm f}(d,t_2)$ if for each
$\langle c, t_1 \rangle \in K^{\rm f}(d,t_2)$ there exits
a confirmation set $B(e,t_3)$
such that: $\langle c, t_1 \rangle \in B(e,t_3)$.
\end{definition}

The following example gives an illustration of a component
that is cancelled for some focused conflict set.
\begin{figure}[ht]
\hspace{3.7cm}
\includegraphics[scale=0.8]{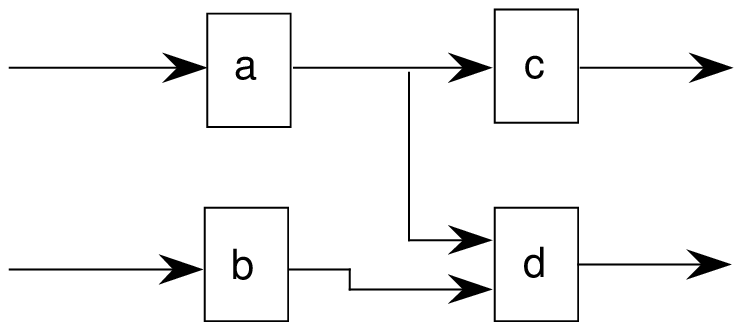}
\caption{} \label{delay}
\end{figure}
\begin{example}
Consider the system in Figure \ref{delay}.
Let component $c$ be the only component depending on a previous
time point and let this component cause a time delay of 1 for its
input value.

Then we can determine the following dependency sets:
\[ {\it Dep}^{\rm f}(c,t) = {\it Dep}^{\rm mf}(c,t) =
\{ \langle a,t-1 \rangle,\langle c,t \rangle \}, \]
\[ {\it Dep}^{\rm f}(c,t+1) = {\it Dep}^{\rm mf}(c,t+1) =
\{ \langle a,t \rangle,\langle c,t+1 \rangle \}, \]
\[ {\it Dep}^{\rm f}(d,t) = {\it Dep}^{\rm mf}(d,t) =
\{ \langle a,t \rangle,\langle b,t \rangle,
\langle d,t \rangle \}, \]
\[ {\it Dep}^{\rm f}(d,t+1) = {\it Dep}^{\rm mf}(d,t+1) =
\{ \langle a,t+1 \rangle,\langle b,t+1 \rangle,
\langle d,t+1 \rangle \}. \]
Suppose that we observe a conflict for output $d$ on time point $t$
and that we observe a confirmation for output $c$ on time point $t+1$.
Then the component $a$ is cancelled for the focused conflict set ${\it Dep}^{\rm f}(d,t)= \{ \langle a,t \rangle,\langle b,t \rangle,
\langle d,t \rangle \}$.
\end{example}
\begin{ruledef} \label{r3}
Given the components in a focused conflict set, focus on the
components that are not cancelled for this conflict set and that are a member of the largest number of other focused
conflict sets .
\end{ruledef}

In a similar way we can also modify Rule \ref{r2} to handle the dynamic
systems in which one fault can be compensated by other faults.
\begin{ruledef} \label{r4}
Given the components in a focused conflict set, focus on the
components that maximise difference between the  number of focused conflict sets of which they are a member and the
number of confirmation sets which cancel the components
for this focused conflict set.
\end{ruledef}

When we have determined the set of focuses of a dynamic
system, we can again determine the most informative
probing point in a focus.
However, to calculate the entropy of a candidate probing point,
we must use the dependency set of this
probing point on the time point in which we want to make the
measurement.

\section{Focusing in systems with loops} \label{loops}
Diagnosis of systems containing loops is a difficult task.
In this section it is shown how we can deal with this task.
The method proposed in not limited to the diagnostic process
described in this paper, but can also be applied in GDE
\cite{Kle-87}.
The problem with loops is that the state of a loop need not be
a function of its inputs only.
It can also depend on the loop it self, in particular on its
previous state if it is a part of a dynamic system.

If a loops is a part of a static system or if it is a part of a
dynamic system whose previous states are unknown,
loops can be dealt with by introducing assumptions \cite{Sta-77}.
When predicting the output value of a component that is part
of a loop, assumptions need to be introduced for those inputs of
the component that are unknown and part of the loop.
Using these assumptions, the output value can be predicted
and propagated.
For each possible value of such an input an assumption has to
be made.

For continues domains it will not be possible to make
assumptions for all possible values.
In that case we need to replace the domain by a finite number
of abstract values.
Fuzzy number, for example, can be one option.

Since the predicted value depends on the assumptions made,
the assumptions have to be stored in the dependency sets.
So, a dependency set contains the assumptions and the
components that must have functioned correctly to observe
the predicted value.

Because assumption are made for inputs that are part of
a loop, after propagating the predictions through the loop,
a prediction will be made for the assumed inputs.
This prediction can either confirm or conflict with the
assumed input value.
If the assumed input value is confirmed, the mask-free
dependency set for the predicted value will be a confirmation
set otherwise its focused dependency set will be a
focused conflict set.
Clearly, if it is a focused conflict set, assuming that non of the
components are broken, the assumption must be wrong.

To focus on the most likely broken components, again
we can apply one of the Rules \ref{r1}, \ref{r2}, \ref{r3} or \ref{r4}.
Here, however, we must take into account the confirmation
sets and the focused conflict sets that arise because of
assumptions that respectively confirm or conflict with the
predicted value.
\begin{figure}[ht]
\hspace{2.8cm}
\includegraphics[scale=0.8]{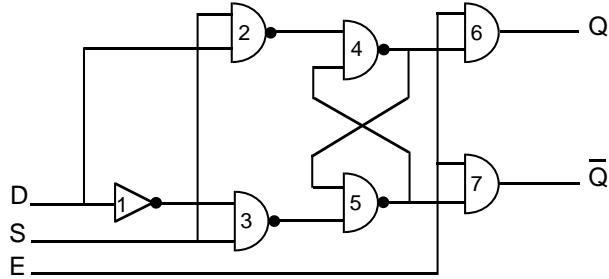}
\caption{flipflop} \label{flipflop2}
\end{figure}
\begin{example}
Consider the flipflop in figure \ref{flipflop2} and
let $D=0$, $S=0$ and $E=1$.
After predicting the output values of each component, the following output
value with their corresponding focused dependency sets
have been derived.
Notice that for digital circuits a mask-free dependency set
is equal to a focused dependency set.
\begin{tabbing}
mmm\=com\=ponentmm\=output\=mvaluemm\=mm\=\kill
\>component\>\>output value\>\>focused dependency set \\
\>\>$inv_1$ \>\>1 \> $\{inv_1\} $ \\
\>\>$nand_2$ \>\>1 \> $\{nand_2\} $ \\
\>\>$nand_3$ \>\>1 \> $\{nand_3\} $ \\
\>\>$nand_4$ \>\>1 \> $\{output(nand_5) = 0,nand_4\} $ \\
\>\>\>\>0 \> $\{output(nand_5) = 1,nand_2,nand_4\} $ \\
\>\>$nand_5$ \>\>0 \> $\{output(nand_5) = 0\}$ \\
\>\>\>\>1 \> $\{output(nand_5) = 1\}$ \\
\>\>$and_6$ \>\>1 \> $\{output(nand_5) = 0,nand_4,and_6\} $ \\
\>\>\>\>0 \> $\{output(nand_5) = 1,nand_2,nand_4,,and_6\} $ \\
\>\>$and_7$ \>\>0 \> $\{output(nand_5) = 0,and_7\} $ \\
\>\>\>\>1 \> $\{output(nand_5) = 1,and_7\} $ \\
\end{tabbing}
Suppose that for both outputs $Q$ and $\overline{Q}$ are equal to $0$.
Then we get the following conflict and confirmation sets
respectively; {\em conflict sets}:
$\{output(nand_5) = 0,nand_4,and_6\} $,
$\{output(nand_5) = 1,and_7\} $ and {\em confirmation sets}: \\
$\{output(nand_5) = 1,nand_2,nand_4,,and_6\} $,
$\{output(nand_5) = 0,and_7\} $.
By applying Rule \ref{r2} we get the focuses: $\{output(nand_5) = 0 \}$,
$\{output(nand_5) = 1 \}$.
Hence, firstly we should measure the output of $nand_5$
to verify the assumptions made.
If, for example, its output value is $1$, we get a new focus:
$\{ and_7 \}$.
\end{example}

The assumptions introduced above for dealing with loops can also
be used for system inputs whose values are unknown.
Unlike loops, here we do not have to consider conflicts between
an assumed value for an input and its predicted value.
A third area where assumptions can be used is in dynamic
systems containing {\em integrating} components.
For example, a capacitor in electrical circuits.
Integrating components can sum up small and irrelevant
differences between their predicted and the actual input values
over time.
After some period of time this can result in a conflict for
some measured output value.
By introducing assumptions around the predicted
output value of the integrators in the focus, we can verify
whether the conflict is caused by small simulation errors.

Above a method for dealing with loops in static systems
and in dynamic systems whose previous states are unknown
is described.
This method opens a loop by introducing assumptions.
In a dynamic system in which we know the previous state of a
loop, there is no need for using assumptions.
Here, we can predict the new state of a loop given the previous
state and the new values for the inputs of the loop provided
that the loop behaves in a deterministic way.

\begin{figure}[ht]
\hspace{1cm}
\includegraphics[scale=0.7]{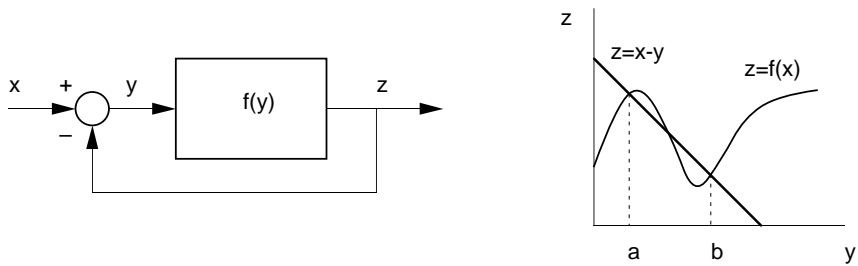}
\caption{algebraic loop} \label{algebra}
\end{figure}

The difficulty of predicting the behaviour of a loop does
not arise because some elements in the loop possess
memory, but because the loop itself can possess memory.
The most well known example of such a loop in a {\em flipflop}
(Figure \ref{flipflop2}).
These loops, depending on the values of their inputs,
can possess multiple equilibrium states.
In the flipflop of Figure \ref{flipflop2}, multiple equilibrium
states arise because the components in the loop function as
switches.
However, also in algebraic loops we can have multiple
equilibrium states.
The loop in Figure \ref{algebra} a gives an illustration.
Here the values $a$ and $b$ are stable solutions of the loop.
Which of the two values we will measure on the system output $z$
depends on the previous state of the system.

Because the state of a loop on a time point $t_i$ depends on
the previous time point $t_{i-1}$, the state of a loop is a function of
its inputs on time point $t$ and the state of the loop in the
previous time point $t_{i-1}$.
When the input value of a component in a loop changes,
and when this input is not a part of the loop, this change
will not have influenced yet the values of the inputs of
this component that are a part of the loop.
This implies that if in the process of predicting the output
values of components we reaches
a component in a loop, we should predict the output value of
this component using for the inputs that are a part of the loop,
the values predicted for in the previous state.
Next the predicted output value should be used to predict the output
values of the other components in the loop.
This can result in a new predicted value for the inputs for which
we have used the values of the previous state.
Even if the predicted value does not change, its dependency
set will change.
These new predicted values should be used to predict again the output
values of the components in the loop.
This prediction process looping through a loop is repeated till
for all the components we have predicted a stable output value
and a stable dependency set.
If the system is well designed such a stable solution must exist.

There is one remark that should be made here.
The predictions process described here results only in a
correct prediction if the system behaves in a deterministic way.
If it does not, the predicted values will depend on the order in which the predictions
are made.

\section{Related work}
\subsubsection*{\it Davis's work}
The focusing method described in this paper can be viewed as a
generalisation Davis's work \cite{Dav-84}.
He uses the {\em single initial cause heuristic} and
the set of causal ancestors of system outputs
for which a conflict was observed, to determine exactly one
focus on likely broken components.

In this paper, instead of sets of causal ancestors, dependency
sets are used.
A dependency set is a set of causal ancestors from which
those components have been removed which do not influence
the predicted output value.
Furthermore, in this paper heuristic rules are used instead of
the single initial cause heuristic.
These rules allow us to determine sets of focuses.
Here, each focus {\em probably} contains one broken component.

After Davis has determined a focus using his single initial
cause heuristic, he eliminates components from the focus
using {\em constraint suspension}.
In principle constraint suspensions can be used in combination
with the diagnostic process described in this paper.
However, constraint suspension requires backward
propagation of predictions, which has a worst case time
complexity that is exponential.

\subsubsection*{\it GDE}
In \cite{Kle-87}, de Kleer and Williams describe a General
Diagnostic Engine.
This GDE derives inconsistencies by forward and backward
propagation of predictions and observations.
Dependencies between predictions made are stored in an ATMS
enabling us to determine the minimal sets of components
involved in the inconsistencies.
These sets of components, which they call conflicts, but which
will be denoted as {\em GDE conflict set} in this section, are
used to determine the set of minimal candidate diagnosis.

\begin{figure}[ht]
\hspace{3.3cm}
\includegraphics[scale=0.8]{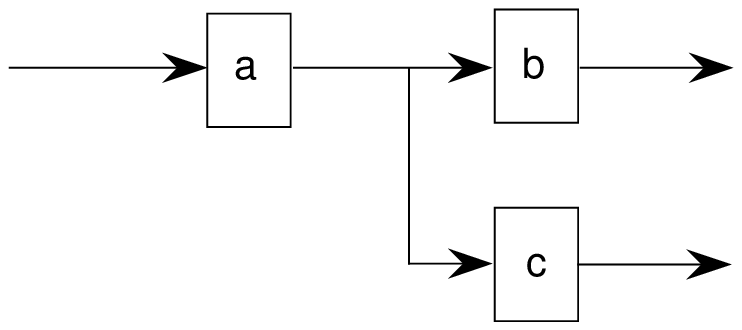}
\caption{} \label{more-less}
\end{figure}

The diagnostic process described in this paper only uses forward
propagation.
Because GDE uses both forward and backward propagation, in some
cases it can gain additional information which lead to a
better set of candidate diagnoses.
To illustrate this consider Figure \ref{more-less}.

Suppose that both the components $b$ and $c$ add $1$ to their in input value.
If we observe  a conflict for the outputs of $b$ and $c$ and if
their output values differ, then in GDE we have
one candidate diagnoses $\{ b,c \}$.
The diagnostic process described in this paper will not
recognise that both $b$
and $c$ must be broken.
Instead it will focus on $\{ a \}$.

Though GDE can give a better diagnosis, as is shown above, this
will in general be insignificant.
Because we assumed that component fail independently, a
situation as is shown in the example above will be very rare.
Especially, when the diagnostic system is monitoring the
system to be diagnosed.
In the example either $b$ or $c$ will break down first.
Hence, if $b$ breaks down first, then, at the time that both
$b$ and $c$ are broken, we will
have determined two focuses; $\{ b \}$ and $\{ a,c \}$.

The diagnostic process described in this paper can give better set
of candidate diagnoses than
GDE, when in GDE backward propagation does
not give relevant additional information.
I.e. the minimal GDE conflict sets are completely determined by
the GDE conflict sets that where derived by forward propagation
of predictions.
In that case the focused conflict set determined by
the diagnostic process described in this paper are minimal
with respect to the GDE conflict sets.
Since here also confirmation sets are used, we get a better
focus on the most likely broken components.
To illustrate this consider again Figure \ref{more-less}.

Suppose that we have observed a conflict for the output of
component $b$ and a confirmation for the output of component
$c$.
Then the focus will contain component $b$, implying
one candidate diagnosis $b$.
Because GDE does not use confirmation sets, it will have two candidate diagnoses $a$ and $b$.
\begin{figure}[ht]
\hspace{2.7cm}
\includegraphics[scale=1.1]{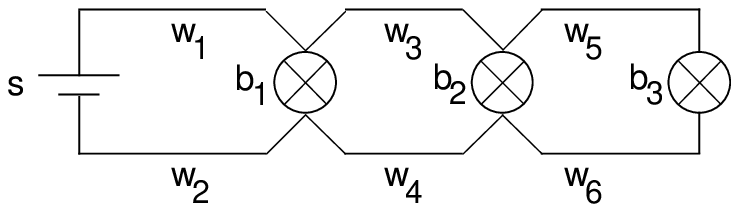}
\caption{} \label{bulb}
\end{figure}

To illustrate the advantage of the use of confirmation sets
more clearly, consider the circuit in Figure \ref{bulb} consisting
of a power supply and three bulbs.
If only bulb 3 gives light, GDE will derive 22 minimal diagnoses.
Among these is the diagnosis in which both bulb 1 and 2 are
broken.
The other diagnoses imply faults like a bulb that is giving light
without power supply, wires that generate energy out of nothing.
Clearly, the diagnosis stating that both bulb 1 and 2 are
broken is the only plausible diagnosis.
To eliminate the implausible diagnoses, fault models
\cite{Kle-89,Str-89} or physical impossibility \cite{Fri-90}
can be used.
Both solutions require additional knowledge.
By using confirmation sets the implausible diagnoses are
eliminate without additional knowledge.
For example, the diagnosis in which bulb 3 is giving light without
power supply is eliminated because bulb 3 giving light is
explained by a power supply that works correctly.

Another difference between the diagnostic process described
in this paper and GDE is the way a new probing point
is selected.
In GDE a probing point is selected by minimising
the expected entropy of the candidate diagnoses after
measuring a probing point.
This implies that every possible outcome of a measurement has
to be considered.
Since the number of candidate diagnoses can grow exponential,
minimising the expected entropy can be every inefficient
unless approximations are used \cite{Kle-91}.

In this paper the amount of information gained by measuring
a probing point is maximised.
As was shown in Section \ref{entropy}, this can be done
efficiently and without approximations.
Furthermore, here possible outcomes of a measurement need not
to be considered.
Hence, the method can also be applied when probing points
take their values from a continuous domain.

\subsubsection*{\it A spectrum of logical definitions}
In \cite{Con-91}, Console and Torasso present a unified
framework for different definitions of Model Based Diagnosis.
This framework integrates abductive and consistency based diagnosis.
It is shown that both abductive and consistency based diagnosis
can be viewed as different points in the spectrum of logical
definition that are introduced by the framework.
More precisely, if no observation has to be explained (abductively),
but all assumptions must be consistent with the observations, then
we are at the bottom of the spectrum.
This corresponds with diagnostic process described by
Reiter \cite{Rei-87}, by
de Kleer and Williams \cite{Kle-87,Kle-89}
and by Struss and Dressler \cite{Str-89}.
At the other end of the spectrum the assumptions must not only be consistent with the observation, but must also
explain the observations.
In between we have the situation in which either the
{\em normal} observations (the observation that correspond
with the predicted values), or the {\em abnormal} observations
must be explained.
The latter, for example, is being applied in  medical diagnosis.

The diagnostic process described in this paper corresponds with
the situation in which only the normal observations (the
observations that confirm the predictions) are explained.
So the diagnoses that can be generated by the set of focuses
are the most likely diagnosis that are consistent with the
observations and that explain the normal observations.

\subsubsection*{\it Non-intermittent faults}
In \cite{Rai-91} Raiman, de Kleer, Saraswat and Shirley describe
how a non-intermittency assumption can be used to improve
the diagnostic precision.
A similar assumption is used in the definition of a component
that is cancelled for a focus conflict set.
The difference between the two approaches is, however, that
Raiman et al. only consider a set of static diagnostic instances
of a system.
They exclude dynamic systems by defining a non-intermittent
behaviour of a component as a behaviour where the outputs
are a function of the inputs.
This definition does not allow for components that possess an
internal state.
Therefore, it excludes dynamic behaviour.
In this paper an intermitting fault is defined as a fault that does
not always result in a faulty output value.
Such a behaviour can arise, for example, because an and-gate in
a digital circuit has a struck at 1 error but also because a
broken component does not always function incorrectly.
External influences, such as the temperature, can cause the latter
kind of intermittency.

\subsubsection*{\it Efficiency}
In general, the goal of model based diagnosis is formulated as
the determination of a diagnosis with the least number of
additional measurements.
Therefore, we try to use as much information as possible.
Also if we are not able to make additional measurements,
to reduce the number of candidate diagnoses, as much
information as possible should be used.
Unfortunately, this can result in a exponential time complexity
for the diagnostic process.
De Kleer, for example, identifies three source for the exponential
time complexity in GDE \cite{Kle-91}; the prediction process,
the conflict recognition and generation of candidate diagnoses.

To avoid these time complexity problems several solution have
been proposed.
All solutions proposed are based on ignoring some of the
information available.
They exchange diagnostic precision for tractability.
In \cite{Kle-91} de Kleer proposes to focus
on likely candidate diagnoses only.
Since we need to know the candidate diagnoses before we can
determine their likelihood, their a priori probabilities are used
as an estimate.
In \cite{Bak-92} Bakker and Bourseau propose a diagnostic
process PDE with a time complexity in ${\cal O}(n^2)$.
They gain their efficiency by reducing the number of conflicts
generated to {\em obvious} and {\em semi-obvious} conflicts and
by determining a probe advice using these conflicts only.
Because a measured output of the system to be diagnosed
can cause an exponential number of conflict sets, as is shown in
Figure \ref{bakker}, they allow only two conflict sets to be
determined for each measured output.
(Their obvious conflicts correspond with the conflict sets
described in this paper.)

Also the diagnostic process described in this paper exchanges
diagnostic precision for tractability.
However, it also uses information that is ignored by
other diagnostic systems.
In some cases this results in better diagnostic precision
GDE or PDE.

\section{Conclusion}
In this paper an efficient diagnostic process is described.
This diagnostic process can be viewed as a generalisation
of Davis's work \cite{Dav-84}.
Furthermore, using the spectrum of logical definitions of
Console and Torasso \cite{Con-91}, the diagnostic process
can be classified as one that determines the most likely
diagnoses that are consistent with the
observations and that explain the normal observations.
Because it explains the normal observation, it will not consider
implausible diagnoses which GDE  will consider when no
fault models \cite{Str-89} or physical impossibility
\cite{Fri-90} is used.

The process described determines a set of focuses on likely
broken components.
These focuses can be viewed as generators of likely candidate
diagnoses.
To refine the set of candidate diagnoses, here an output value of
a component in a focus must be measured.
The best output to be measured, is selected using the entropy
of each possible probing point in the focus.
It is shown that we can determine the entropy of a candidate
probing point
only using the focus and the set of components on which a predicted value of the probing point
actually depends.
Because it is not necessary to consider the possible values of a
candidate probing point, this method is also applicable when
the domain of values for a probing point is continuous.

It is shown how the diagnostic process can also be applied
for dynamic systems and systems containing loops.
When diagnosing dynamic systems their is no need to demand
that components do not posses an internal state.
Furthermore, depending on whether we make a non-intemittency
assumption, a better set of focuses can be determined.

A important advantages of the diagnostic process
described in this paper is its efficiency.
The diagnostic process has a worst case complexity of
${\cal O}(n^2)$ where $n$ is the number of components.
However, if the connectivity between the components is low,
the average time complexity becomes linear.

Because of its low time complexity, the diagnostic process
described in this paper is especially suited for diagnosing
dynamic systems.
In such systems, typically, the outputs of a specific set of
components are sampled.
Here, a diagnostic process must be efficiently enough
to process the information gained from a sample before
the next sample is made.

Finally one should notice that the diagnostic process described
can be combined with other technologies like
constraint suspension and abductive diagnosis.
Because each focus probably contains exactly one broken
component, the set of focuses can be used to reduce the
time complexity of these technologies.

\end{document}